\documentclass{article}
\ifdefined\pdfsuppressptexinfo
\fi
\usepackage{iclr2026_conference,times}

\usepackage{amsmath,amsfonts,bm}

\def\eqref#1{equation~\ref{#1}}

\def\1{\bm{1}}

\DeclareMathAlphabet{\mathsfit}{\encodingdefault}{\sfdefault}{m}{sl}
\SetMathAlphabet{\mathsfit}{bold}{\encodingdefault}{\sfdefault}{bx}{n}

\usepackage{booktabs}
\usepackage{graphicx}
\usepackage{float}
\usepackage{array}
\usepackage{microtype}
\usepackage{hyperref}
\usepackage{xurl}
\hypersetup{hidelinks}

\title{IndustrialCAD-1400: Benchmarking\\Detailed Industrial Assemblies from Text}

\author{%
\begin{minipage}{\dimexpr\textwidth-2\tabcolsep\relax}
\centering
{\large\bfseries
Hebao Zhu$^{2,*}$\quad Kang An$^{1,*,\dagger}$\quad Zhuoke Liao$^{2}$\quad FaQiang Qian$^{3}$\\
Ziliang Wang$^{3}$\quad Mingang Chen$^{4}$\quad Qibing Ren$^{1,\ddagger}$\\[0.4em]}
{\normalfont\normalsize
$^{1}$Shanghai Jiao Tong University\quad $^{2}$Chongqing University\\
$^{3}$SenseTime\\
$^{4}$Shanghai Development Center of Computer Software Technology\\[0.3em]}
{\normalfont\small\ttfamily
20231008@stu.cqu.edu.cn\quad \{an\_kang, renqibing\}@sjtu.edu.cn}\\[0.3em]
{\normalfont\footnotesize
$^{*}$Equal contribution.\quad $^{\dagger}$Project leader.\quad $^{\ddagger}$Corresponding author.}
\end{minipage}%
}

\newcommand{\benchname}{IndustrialCAD-1400}

\iclrfinalcopy
\begin{document}
\maketitle

\begin{abstract}
Industrial assemblies require more than a recognizable shape: functional parts, interfaces, dimensions and service features must satisfy a detailed request together. 
We introduce IndustrialCAD-1400, a text-to-CAD benchmark for this industrial setting. Its 1,400 requests combine 400 equipment categories across 40 domains with 250 four-case design families. 
Family requests specify a median of 17 named component or configuration roles and six explicit numeric targets; controlled variants test whether dimensions change while shared structure is retained. 
Task construction draws on documented manufacturer information, owner specifications and engineering guidance. 
Models generate geometry programs from requests, and a common Blender environment executes them to produce scenes, optional measurements, and standardized views for evaluation. 
A request assessment of front, isometric, and top views measures visible component, arrangement, and interface fulfillment without requiring a single reference geometry. 
Five stratified panels preserve complete families while covering the benchmark exactly once, enabling balanced comparisons and separating industrial breadth from controlled numerical responsiveness. 
In conclusion, IndustrialCAD-1400 extends text-to-CAD from recognizable part geometry to industrial assembly synthesis under joint component, interface, spatial, and numerical specifications.
\end{abstract}

\section{Introduction}

Generating industrial assemblies from natural language requires coordinating component roles, dimensions, and spatial relationships within a common design.
Advances in geometric representations, construction-sequence modeling, and language-conditioned generation have established foundations for this capability
\citep{koch2019abc,willis2021fusion,wu2021deepcad,khan2024text2cad}.
Industrial evaluation must examine whether these capabilities extend across equipment types and satisfy multiple assembly requirements simultaneously.

\begin{figure}[t]
\centering
\includegraphics[width=\linewidth]{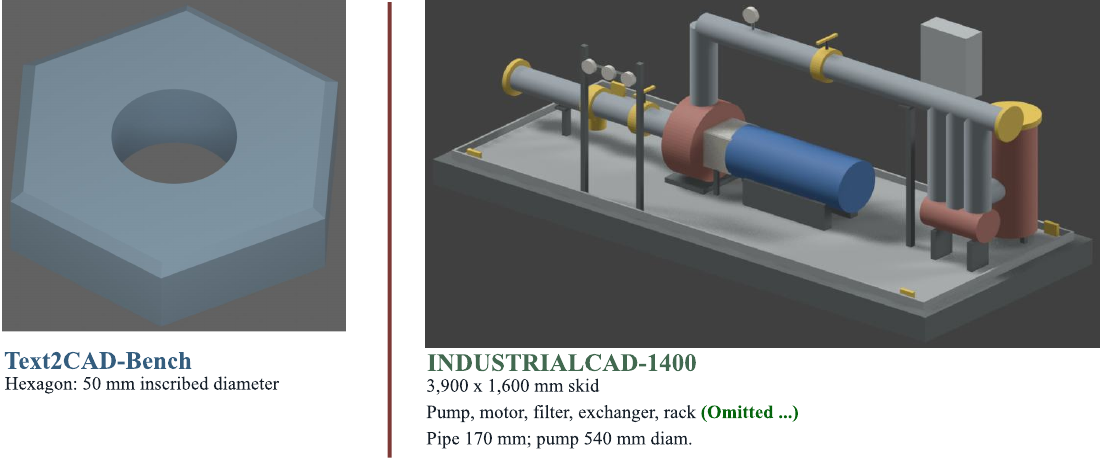}
\caption{Part-level and assembly-level modeling requirements. Left: a single-part
example rendered from the published L1 reference code in Text2CAD-Bench
\citep[Appendix F]{wang2026text2cad}. Right: a pump-skid reference assembly
from \benchname{}. The examples illustrate local geometric features and
coordinated component, layout, and dimensional requirements, respectively.
The complete family-request structure appears in Figure~\ref{fig:complete-task-spec}.}
\label{fig:request-comparison}
\end{figure}

Existing benchmarks address complementary modeling scopes. Text2CAD-Bench broadens part-level evaluation while restricting its application tier to single-body designs without assembly or mating constraints
\citep{wang2026text2cad}.
Assembly research such as JoinABLe instead predicts relationships between supplied parts
\citep{willis2022joinable}.
A pump-skid request instead requires generating a pump, motor, filter vessel, pipework, and frame that jointly satisfy component, layout, and dimensional requirements.
Figure~\ref{fig:request-comparison} illustrates this distinction in modeling scope.

Two evaluation objectives follow: representing characteristic equipment structures across domains, and realizing numerical changes within a shared assembly configuration.
Broad category coverage alone does not isolate parameter responsiveness; variants within a few configurations do not establish industrial breadth.
Separate, complementary tasks allow both capabilities to be studied explicitly.

Assessment must likewise account for both measurable geometry and visible assembly fidelity.
Executability alone does not establish specification compliance, and similarity to a single reference can penalize alternative valid designs.
CADTests advances requirement-based evaluation through executable geometric and topological checks
\citep{mallis2026text}.
For industrial assemblies, these checks can be complemented by visual assessment of component form, layout, and configuration.
Balanced comparisons also require task selection that preserves complete design families, motivating a stratified evaluation design alongside geometric and image-based assessment.

\begin{figure}[t]
\centering
\includegraphics[width=\linewidth]{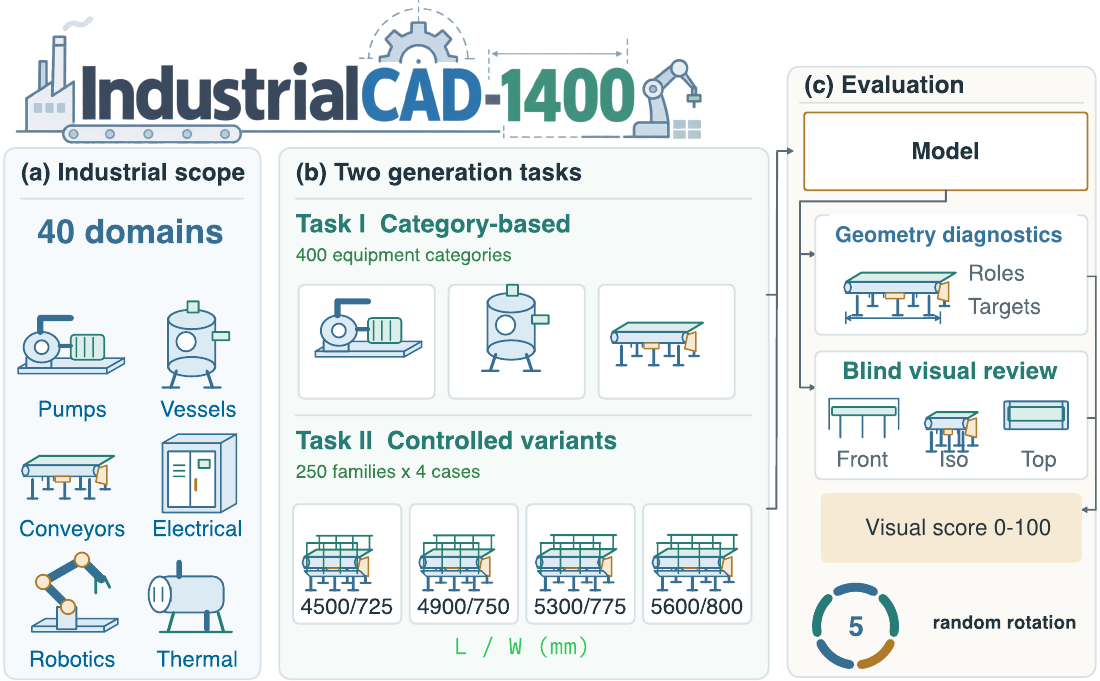}
\caption{Overview of \benchname{}. Industrial coverage motivates category-based
and controlled-variant generation. Blind three-view review provides the shared
visual-conformity score; geometric diagnostics denote optional measurement extensions.
Five disjoint, stratified panels cover all requests once. Equipment and views are
schematic; the four conveyor dimension pairs are actual task targets.}
\label{fig:benchmark-overview}
\end{figure}

We introduce \benchname{}, whose requests draw on industrial documentation and allow multiple valid geometric realizations of a specification.
Figure~\ref{fig:benchmark-overview} summarizes the benchmark design.
Our contributions are threefold:
\begin{itemize}
    \item \textbf{An industrial benchmark with broad equipment coverage.}
    We introduce 1,400 requests spanning 40 industrial domains, including process equipment, material handling, electrical systems, robotics, and thermal management.
    The benchmark makes equipment-specific components and configurations central to text-to-CAD evaluation.

    \item \textbf{Two complementary industrial generation tasks.}
    Category-based generation covers 400 equipment types, while controlled-variant generation comprises 250 four-case assembly families.
    Together, the tasks assess industrial breadth and numerical responsiveness under shared structural requirements.

    \item \textbf{Visual evaluation with randomized rotation.}
    We assess external assembly conformity with one shared visual standard across all tasks, using blind review of three standardized views; geometric measurement is treated as a separate diagnostic extension.
    A stratified randomized rotation protocol provides five balanced, disjoint evaluation panels, preserving complete families and supporting matched comparisons at both partial and full benchmark budgets.
\end{itemize}

\section{Related Work}

\paragraph{Representations and CAD construction.}
ABC supplies boundary representations; Fusion 360 Gallery provides
construction sequences \citep{koch2019abc,willis2021fusion}.
SketchGraphs models primitive-and-constraint graphs
\citep{seff2020sketchgraphs}. DeepCAD and SkexGen model sketch-and-extrude
programs \citep{wu2021deepcad,xu2022skexgen}; BRepNet, SolidGen, and BrepGen
model boundary representations
\citep{lambourne2021brepnet,jayaraman2023solidgen,xu2024brepgen}.
CAD-Recode reconstructs executable CadQuery programs from point clouds
\citep{rukhovich2025cad}. Our benchmark evaluates industrial specification fulfillment.

\paragraph{Language-guided and interactive generation.}
Text2CAD generates CAD sequences from text
\citep{khan2024text2cad}. CADFusion incorporates visual feedback during training
\citep{wang2025cadfusion}; CAD-Assistant uses FreeCAD tools for iterative
multimodal modeling \citep{mallis2025cadassistant}. CADCodeVerify studies
vision-language code generation and refinement
\citep{alrashedy2025generating}. We evaluate specification fulfillment under
common execution conditions with one bounded opportunity to correct an execution
error, retaining initial and final outcomes separately.

\paragraph{Assemblies and requirement-based benchmarks.}
JoinABLe predicts parametric joints between supplied parts
\citep{willis2022joinable}, whereas our tasks require generating the components
and their layout from text. Text2CAD-Bench provides 600 examples in four
complexity tiers with paired prompt styles; its application tier is restricted
to single-body designs
\citep{wang2026text2cad}. CADTests evaluates prompt-specified properties
using tests validated by soundness and mutation analysis
\citep{mallis2026text}. We build on requirement-level evaluation with two
industrial tasks that distinguish equipment coverage from controlled assembly variation.

\paragraph{Visual assessment and judge reliability.}
TIFA and VQAScore assess text--image alignment
\citep{hu2023tifa,lin2025vqascore}. LLM judges can agree with human ratings
while retaining systematic biases \citep{zheng2023judging}. We use request-grounded
visual observations under a shared 0--100 scale, retain geometric measurements
as diagnostics, and specify an independent human-agreement study; shared
rotation panels control task selection.

\section{Benchmark}
\label{sec:benchmark}

\benchname{} comprises 1,400 requests across 40 industrial domains,
organized to assess equipment coverage and controlled specification variation.

\subsection{Industrial Scope and Task Construction}

Industrial documentation informs the components, interfaces, and configuration
features represented in the benchmark. Sources include manufacturer and supplier
information, owner specifications, and engineering guidance.
Table~\ref{tab:provider-documents} illustrates their role in task construction:
vessel specifications motivate nozzles and access features, while conveyor
guidance informs pulleys, tensioners, and maintenance-related components.
Public source labels connect requests to supporting specifications and domain notes.

\begin{table}[t]
\centering\small
\caption{Industrial documentation informing task construction. Source names and
descriptions appear in Appendix Table~\ref{tab:source-examples}.
Numerical task targets are defined by the benchmark.}
\label{tab:provider-documents}
\begin{tabular}{@{}p{0.22\linewidth}p{0.34\linewidth}p{0.37\linewidth}@{}}
\toprule
Provider & Document & Task-construction context \\
\midrule
ADNOC & Pressure Vessel Specification & Nozzles, manways, supports and access \\
Flexco & Belt Conveyor Products handbook & Belt accessories, pulleys and tensioners \\
NSW government & MDG 4 conveyor-tensioning guideline & Tensioning and maintenance context \\
\bottomrule
\end{tabular}
\end{table}

The requests specify equipment identity, components, and configurations, with
numerical targets in the family task. Appendix~\ref{app:source-provenance}
details source coverage and provenance limits.

\subsection{Two Complementary Generation Tasks}
\label{sec:tasks}

Table~\ref{tab:data-profile} summarizes the two tasks. Roles denote named component
or configuration targets, rather than physical solids; numerical items count
explicit scalar or count requirements, excluding metadata and evaluation tolerances.

\begin{table}[t]
\caption{Profiles of the two generation tasks. Roles and numerical items are per-request medians [min,max], counting named component/configuration targets and explicit numerical requirements, respectively. Enhanced lists requests/additional checks (136/707 overall).}
\label{tab:data-profile}
\centering
\small\fontsize{9}{11}\selectfont
\setlength{\tabcolsep}{4pt}
\begin{tabular}{@{}lrrlccc@{}}
\toprule
Track & $N$ & Domains & Organization & Roles & Numeric & Enhanced \\
\midrule
Taxonomy & 400 & 40 & 10 per domain & 4 [3,5] & 0 [0,0] & 45/228 \\
Family variants & 1,000 & 5 & 250 families $\times$ 4 & 17 [10,22] & 6 [5,8] & 91/479 \\
\bottomrule
\end{tabular}
\end{table}

\paragraph{Task I: Category-based generation.}
The taxonomy track evaluates whether a model can generate the characteristic
components and configuration of diverse industrial equipment. It contains
400 requests, with ten equipment categories in each of 40 domains. Each request
specifies an equipment type and its required roles, with a median of four named
component or configuration targets. Evaluation emphasizes component completeness
and visible organization across domains. The task has no accepted reference
geometry and assesses conformity to the request rather than shape matching.

\begin{figure}[t]
\centering
\includegraphics[width=\linewidth]{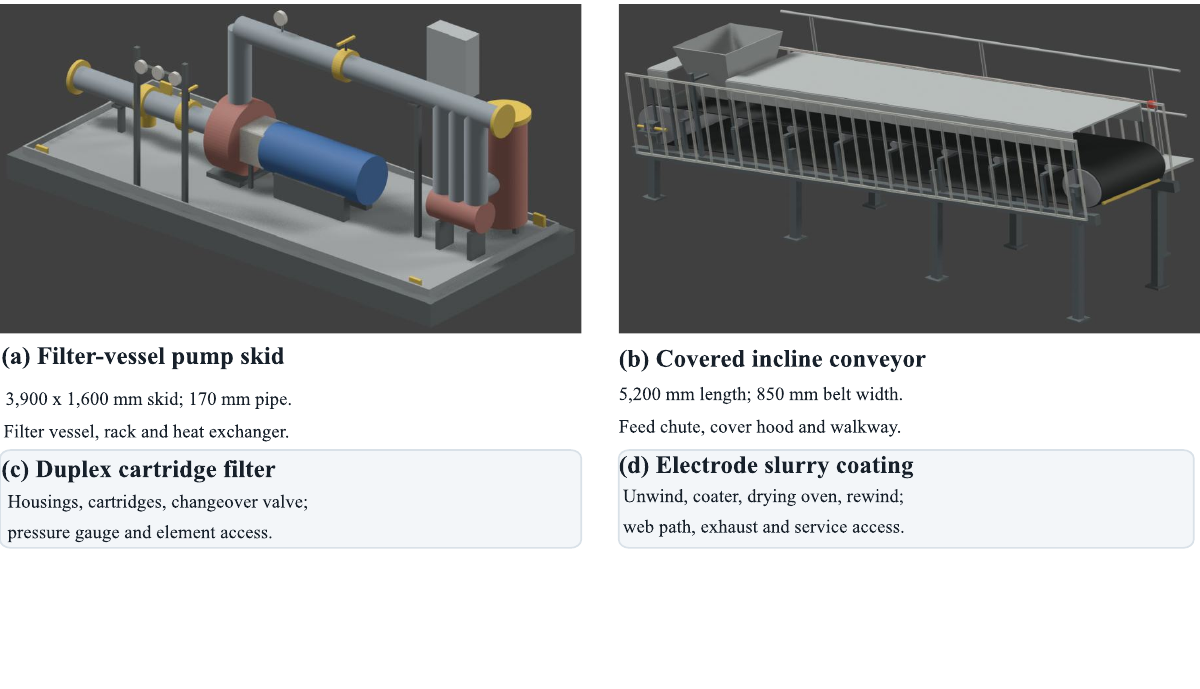}
\vspace{-15mm}
\caption{Examples from the two generation tasks. Controlled-variant examples
(a,b) show reference assemblies for a pump skid and a covered conveyor, together
with component and numerical requirements. Category-based examples (c,d)
specify equipment identity and required features without reference geometry.}
\label{fig:sample-gallery}
\end{figure}

\paragraph{Task II: Controlled-variant generation.}
The family track evaluates whether numerical specifications are realized
consistently within detailed assemblies. It contains 1,000 requests organized
into 250 four-case families. Five equipment groups contribute 200 cases each:
pump skids, belt conveyors, electrical enclosures, cooling towers, and pressure
vessels. Within a family, numerical targets vary while topology and material
specifications remain fixed. This design tests both parameter responsiveness
and preservation of shared components and organization. Family requests contain
a median of 17 named roles and six numerical targets.
Figure~\ref{fig:sample-gallery} illustrates both tasks.

\subsection{Task Interface and Requirement Coverage}

\begin{figure}[t]
\centering
\includegraphics[width=0.96\linewidth]{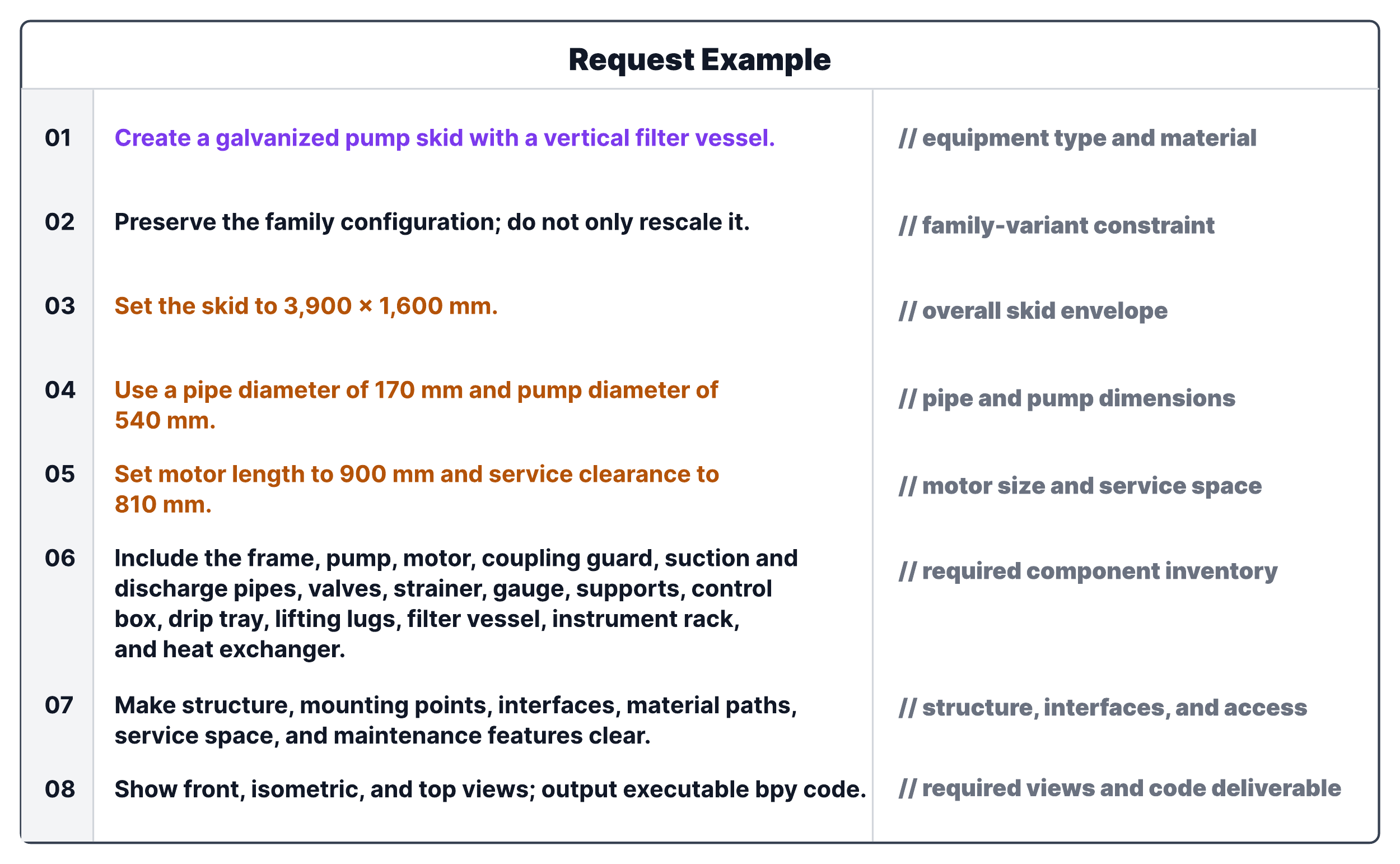}
\caption{Representative family request example. 
The request fixes the equipment identity, numerical targets, named component roles,
and three-view evidence requirements while allowing multiple valid geometries.
}
\label{fig:complete-task-spec}
\end{figure}

Both tasks map a natural-language request to a Python geometry program using
Blender operations or shared primitive helpers, with meters and radians as the
common units. The model receives the request and common modeling instructions;
source documents, reference programs, and evaluation measurements are excluded
from its input. Figure~\ref{fig:complete-task-spec} gives a complete
representative family request and its requirement structure. Only requested
components and applicable interfaces enter requirement checks; inherited interface lists provide
conditional guidance rather than additional universal constraints.

A targeted subset makes additional structural requirements explicit in the
model-visible requests: 45 taxonomy and 91 family cases contain 228 and 479
additional checks, respectively. These assess distinct component geometry,
mesh closure, enclosure hollowness and containment, and conveyor placement
relations. The family subset comprises 69 conveyors and 22 enclosures.
Because these cases were selected for their structural requirements, subgroup
results characterize this task subset rather than the effect of random assignment.

\section{Evaluation}
\label{sec:evaluation}

The evaluation measures how well a generated assembly satisfies the visible
requirements of its text specification. It combines standardized image evidence,
a common visual score, and stratified task panels. The same protocol applies to
category-based and controlled-variant generation.

\subsection{Execution and Standardized Views}

Generated programs are executed in a common Blender environment. A valid delivery
must produce a nonempty scene that can be rendered under the evaluation settings.
Each delivered assembly is represented by front, isometric, and top orthographic
views at $1024\times1024$ resolution. Camera directions are fixed across tasks,
while framing adapts to the assembly's extent. Rendering preserves the submitted
geometry; it does not add missing components or repair the assembly.

The three views expose complementary evidence about component form, spatial
layout, and external interfaces. The judge receives the text request and these
views, without the generator's identity or implementation. This permits different
geometric realizations of the same specification to be assessed without matching
a single reference shape. Appendix~\ref{app:execution} gives execution and camera
settings.

\subsection{Score}

For a valid delivery, Evaluator assigns a holistic integer score
$J\in\{0,1,\ldots,100\}$ using one rubric shared by all tasks and models. The rubric
considers whether the requested equipment is recognizable and whether its
visible components, main arrangement, and interfaces satisfy the specification.
The judge records supporting observations, discrepancies, and uncertainties,
linking each to a request excerpt and a location in the supplied views.

Reference anchors at 0, 25, 50, 75, and 100 describe progressively greater
fulfillment, from no meaningful match to a consistent external assembly. They
guide interpretation while permitting every integer score; components and
defect types do not carry fixed point allocations. Thus, the score reflects the
assembly as a whole rather than a sum of task-specific checks.

The metric concerns visible external conformity. 
A normally closed housing does not require an unrequested cutaway,
whereas an unclear required external connection remains relevant to the visual
assessment. Appendix~\ref{app:visual-review} provides the full rubric and evidence
requirements; optional geometric diagnostics are documented in
Appendix~\ref{app:measurement} and do not contribute to the main score.

The reported score includes delivery failures:
\begin{equation}
S=\begin{cases}
J, & \text{valid model delivery},\\
0, & \text{model-output failure after the permitted repair}.
\end{cases}
\label{eq:visual-score}
\end{equation}
Infrastructure and assessment failures are labeled unresolved, not model
zeros. The first valid image assessment is retained; response validation checks
the required fields, score range, and request quotations without changing the
judge's score.

\subsection{Stratified Randomized Rotation}
\label{sec:rotation}

The 1,400 requests are assigned to five fixed, disjoint panels by stratified
random sampling. Each round contains two category requests from each of the
40 industrial domains, giving 80 category requests, and ten complete families
from each of the five variant domains, giving 200 variant requests. All four
cases of a family remain together so that shared structure and requested
changes can be compared within the same round.

All generators use the same panel assignment. Each round therefore compares
models on an identical 280 request with the same domain balance, and the
five rounds cover the benchmark exactly once. Differences between round scores
describe sensitivity to task composition. The partition and random seed are
specified in Appendix~\ref{app:visual-review}.

\subsection{Score Aggregation}

We report category and variant means separately. Category scores are averaged
within each domain and then equally across domains; variant scores are averaged
within each four-case family and then equally across families. The balanced
design makes these equivalent to request-level means within each task. All
scheduled model outputs, including delivery-failure zeros, remain in the
denominator. The overall score for round $r$ is
\begin{equation}
\bar S_r=\frac{2\bar S_{\mathrm{category},r}
                   +5\bar S_{\mathrm{variant},r}}{7}.
\label{eq:round-score}
\end{equation}
Since the panels have equal sizes, the full-benchmark score is
$\bar S_{\mathrm{full}}=\frac{1}{5}\sum_{r=1}^{5}\bar S_r$.

Delivery rate reports the fraction of requests yielding valid outputs. We also
report a conditional mean over delivered outputs to distinguish failures to
produce an assembly from low visual conformity after delivery. 

\section{Experiments}
\label{sec:experiments}

We compare DeepSeek-V4-Flash, DeepSeek-V4-Pro, and MiniMax-M3, with GPT6—Astra
used only for image assessment. All generators share Blender 4.4.1, the modeling
helpers, rendering settings, and visual rubric. Generation uses a 15,000-token
limit per request with reasoning disabled. Each task permits an initial response
and at most one execution-error feedback repair, following the feedback principle
of Text2CAD-Bench~\citep{wang2026text2cad}. The final attempt is evaluated
without selecting the better of the two outputs. These settings hold across
both tasks and all rotation panels. 
In the following text, Pro refers to DeepSeek-V4-Pro, and Flash refers to DeepSeek-V4-Flash.

\subsection{Comparison Across the Two Tasks}

Table~\ref{tab:model-full-evaluation} separates delivery and visual conformity for the two tasks. 
Pro has the highest overall score at 67.63, followed by Flash at 63.08 and MiniMax at 53.02. 
The corresponding delivery rates are 96.43\%, 92.86\%, and 85.00\%. 
Thus, the overall ordering reflects both successful delivery and the conformity of the resulting assemblies.

The two tasks expose different model differences. Category scores occupy a
narrow range, from 54.10 to 56.00, whereas variant scores range from 52.39 to
72.28. Pro and Flash score 16.28 and 12.58 points higher on variants than on
categories; MiniMax scores 2.22 points lower. MiniMax also slightly exceeds Pro
on categories while falling behind it on variants. Reporting only the overall
mean would obscure this difference between broad equipment coverage and
generation within specified assembly families.

\begin{table*}[t]
\caption{Delivery and visual conformity on the two generation tasks. Delivery is
the percentage of valid outputs; Score is the mean 0--100 score including
model-failure zeros. Overall uses 2:5 task weights.}
\label{tab:model-full-evaluation}
\centering\small
\setlength{\tabcolsep}{6pt}
\begin{tabular}{@{}lrrrrrr@{}}
\toprule
 & \multicolumn{2}{c}{Category} & \multicolumn{2}{c}{Variant} & \multicolumn{2}{c}{Overall} \\
\cmidrule(lr){2-3}\cmidrule(lr){4-5}\cmidrule(l){6-7}
Model & D & S & D & S & D & S \\
\midrule
DeepSeek-V4-Pro & 95.00 & \textbf{56.00} & 97.00 & \textbf{72.28} & 96.43 & \textbf{67.63} \\
DeepSeek-V4-Flash & 88.75 & 54.10 & 94.50 & 66.68 & 92.86 & 63.08 \\
GLM-5.3 & 94.00 & 55.30 & 86.00 & 56.14 & 88.29 & 55.90 \\
MiniMax-M3 & 92.50 & 54.60 & 82.00 & 52.39 & 85.00 & 53.02 \\
\bottomrule
\end{tabular}
\end{table*}

Category and variant tasks differ in equipment coverage and specification
structure, so their score difference is not a controlled estimate of the benefit
of more detailed prompts.

\subsection{Performance Across Equipment Domains}

Table~\ref{tab:industrial-detail-results} breaks the variant task into its five
equipment domains, reporting delivery alongside score. Pro leads in score
across all five. Conveyors receive the highest scores for every model: 78.38
for Pro, 76.70 for Flash, and 66.83 for MiniMax. The Pro--Flash gap is only
1.68 points for conveyors, compared with 9.98 points for pressure vessels.
Model differences therefore vary substantially across equipment groups.

\begin{table}[t]
\caption{Controlled-variant results by equipment domain. 
Pro refers to DeepSeek-V4-Pro, and Flash refers to DeepSeek-V4-Flash.
D is delivery rate(\%); S is the mean visual score including model-failure zeros. Domains have
equal weight. Bold denotes the highest score in each row.}
\label{tab:industrial-detail-results}
\centering\small
\setlength{\tabcolsep}{3.5pt}
\begin{tabular}{@{}lrrrrrrrr@{}}
\toprule
 & \multicolumn{2}{c}{Pro} & \multicolumn{2}{c}{Flash} & \multicolumn{2}{c}{GLM-5.3} & \multicolumn{2}{c}{MiniMax-M3} \\
\cmidrule(lr){2-3}\cmidrule(lr){4-5}\cmidrule(lr){6-7}\cmidrule(l){8-9}
Equipment domain & D & S & D & S & D & S & D & S \\
\midrule
Electrical enclosures & 100.00 & \textbf{67.33} & 97.50 & 58.10 & 96.00 & 57.05 & 100.00 & 54.08 \\
Cooling towers & 100.00 & \textbf{68.98} & 100.00 & 63.70 & 91.00 & 56.89 & 87.50 & 54.20 \\
Conveyors & 97.50 & \textbf{78.38} & 97.50 & 76.70 & 90.00 & 68.74 & 85.00 & 66.83 \\
Pressure vessels & 95.00 & \textbf{74.10} & 85.00 & 64.13 & 68.00 & 41.92 & 57.50 & 33.88 \\
Pump skids & 92.50 & \textbf{72.60} & 92.50 & 70.75 & 85.00 & 56.10 & 80.00 & 52.95 \\
\bottomrule
\end{tabular}
\end{table}

Pressure vessels produce the largest Pro--MiniMax gap, at 40.23 points.
MiniMax delivers only 57.50\% of vessel outputs, compared with 95.00\% for Pro
and 85.00\% for Flash, so the vessel score includes a substantial contribution from
delivery failures. By contrast, Pro and MiniMax both deliver every electrical-enclosure
output, yet their scores differ by 13.25 points. This separates two sources
of difficulty: producing a valid assembly and satisfying its visible requirements.
The domain table supports these comparisons without attributing the differences
to unmeasured geometric properties or specific defect types.

\subsection{Score Distribution}

\begin{figure}[t]
\centering
\includegraphics[width=\linewidth]{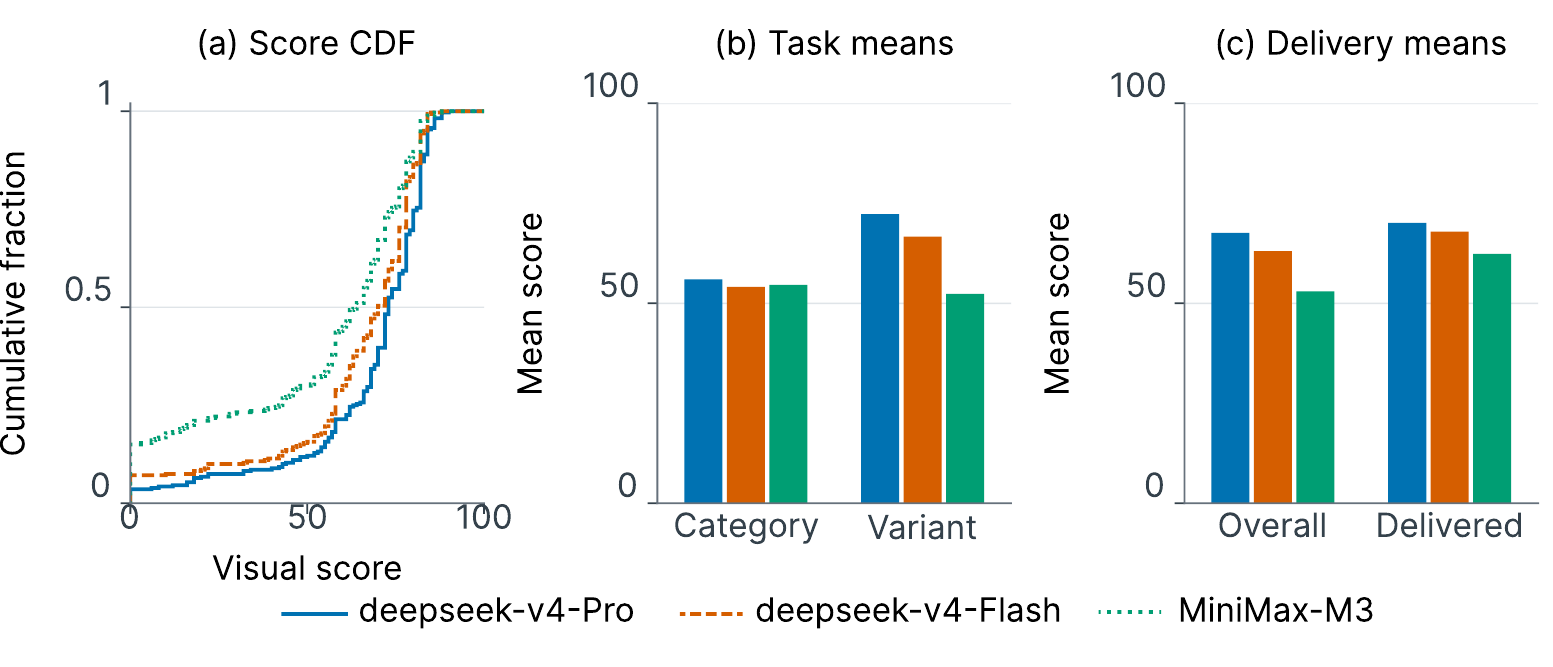}
\caption{Visual conformity across models. (a) Cumulative score distributions;
(b) mean scores for the two tasks; (c) means over all outputs and over valid
deliveries only. All-output means include model-failure zeros. Conditional means
use each model's own delivered subset.}
\label{fig:analysis-preview}
\end{figure}

Figure~\ref{fig:analysis-preview}(a) shows the cumulative score distributions.
At a fixed score, a lower curve indicates fewer outputs at or below that score.
Scores of at least 75 account for 45.36\% of Pro outputs, 38.21\% of Flash outputs,
and 24.64\% of MiniMax outputs. This descriptive threshold illustrates differences
in the upper part of the distribution; it is not a pass criterion.
Figure~\ref{fig:analysis-preview}(b) summarizes the task means, making the larger
separation on variants visible alongside the similar category scores.

Conditioning on delivery raises the means to 70.13, 67.93, and 62.37 for Pro,
Flash, and MiniMax, respectively (Figure~\ref{fig:analysis-preview}(c)). The ordering
is preserved, but MiniMax's mean increases by 9.36 points, compared with 2.50
for Pro and 4.85 for Flash. Delivery failures therefore account for a larger
reduction in MiniMax's overall score. The relation
$\bar S=d\bar J_{\mathrm{delivered}}$, where $d$ is the delivery fraction,
makes this contribution explicit. The conditional means describe each model's
successful outputs.

\subsection{Randomized Rounds}

Table~\ref{tab:randomized-rotation} compares models on the five shared panels
from Section~\ref{sec:rotation}, with separate category, variant, and weighted
overall scores. This layout identifies which task contributes to a model's
round-level changes and whether its ordering relative to other models is
consistent across panels. The fixed domain balance isolates differences among
selected categories and families from changes in domain proportions. The
round-level results are complete and report this variation directly.

\begin{table}[t]
\caption{Completed scores across five shared, disjoint stratified panels, each
containing 80 category and 200 variant requests. Means include model-failure
zeros; Overall uses 2:5 task weights.}
\label{tab:randomized-rotation}
\centering\small
\setlength{\tabcolsep}{4pt}
\begin{tabular}{@{}llrrrrrr@{}}
\toprule
Model & Task & Round 1 & Round 2 & Round 3 & Round 4 & Round 5 & Mean \\
\midrule
DeepSeek-V4-Pro & Category & 53.90 & 57.80 & 55.35 & 58.40 & 54.55 & \textbf{56.00} \\
 & Variant & 69.83 & 74.33 & 71.38 & 75.03 & 70.81 & \textbf{72.28} \\
 & Overall & 65.28 & 69.61 & 66.80 & 70.28 & 66.16 & \textbf{67.63} \\
\addlinespace[3pt]
DeepSeek-V4-Flash & Category & 51.84 & 56.20 & 53.35 & 56.55 & 52.56 & \textbf{54.10} \\
 & Variant & 64.53 & 68.98 & 65.58 & 69.13 & 65.16 & \textbf{66.68} \\
 & Overall & 60.90 & 65.33 & 62.09 & 65.54 & 61.56 & \textbf{63.08} \\
\addlinespace[3pt]
GLM-5.3 & Category & 53.05 & 57.35 & 54.70 & 57.50 & 53.90 & \textbf{55.30} \\
 & Variant & 53.74 & 58.29 & 55.39 & 58.44 & 54.84 & \textbf{56.14} \\
 & Overall & 53.54 & 58.02 & 55.19 & 58.17 & 54.57 & \textbf{55.90} \\
\addlinespace[3pt]
MiniMax-M3 & Category & 52.10 & 56.95 & 53.75 & 56.70 & 53.50 & \textbf{54.60} \\
 & Variant & 49.74 & 54.59 & 51.49 & 54.74 & 51.37 & \textbf{52.39} \\
 & Overall & 50.41 & 55.26 & 52.14 & 55.30 & 51.98 & \textbf{53.02} \\
\bottomrule
\end{tabular}
\end{table}

\subsection{Human--Judge Agreement}

The completed human-agreement study uses the same 80 requests per generator:
one category request per industrial domain and ten complete four-case families.
Two domain-familiar reviewers independently score each valid output from the
request and three views using the common 0--100 rubric, without seeing model
identities or GPT6-Astra assessments.

Table~\ref{tab:human-model-calibration} separates the tasks and their combined
panel. Mean absolute error compares GPT6-Astra with the mean of the two human
scores; Spearman correlation measures agreement in output ordering. Only outputs
with both human ratings and a valid judge assessment enter these statistics;
delivery failures are recorded separately. The completed measurements quantify
agreement between the image judge and domain-informed assessments.

\begin{table}[t]
\caption{Completed human--judge comparison using 80 requests per model and two
independent reviewers. MAE and Spearman correlation compare the judge with the
mean human score.}
\label{tab:human-model-calibration}
\centering\small
\setlength{\tabcolsep}{6pt}
\begin{tabular}{@{}lrrrrrr@{}}
\toprule
 & \multicolumn{2}{c}{Category} & \multicolumn{2}{c}{Variant} & \multicolumn{2}{c}{Overall} \\
\cmidrule(lr){2-3}\cmidrule(lr){4-5}\cmidrule(l){6-7}
Model & MAE & $\rho$ & MAE & $\rho$ & MAE & $\rho$ \\
\midrule
DeepSeek-V4-Pro & 5.41 & 0.923 & 6.36 & 0.872 & 5.89 & 0.924 \\
DeepSeek-V4-Flash & 6.31 & 0.825 & 6.24 & 0.847 & 6.28 & 0.861 \\
GLM-5.3 & 7.09 & 0.877 & 6.90 & 0.884 & 6.99 & 0.889 \\
MiniMax-M3 & 6.95 & 0.893 & 6.49 & 0.799 & 6.72 & 0.865 \\
\bottomrule
\end{tabular}
\end{table}




\section{Conclusion}

\benchname{} provides 1,400 requests across 40 industrial domains, distinguishing
generation across 400 equipment categories from controlled variation within
250 assembly families. A shared visual-conformity standard and stratified randomized
rotation support comparison of industrial coverage and external assembly fidelity,
with separate reporting for category-based and controlled-variant generation.
The comparison reveals larger model differences on controlled variants than on
category-based generation. 
\label{sec:main-end}

\section*{AI Use}
\label{sec:ai-use}
AI assistants supported manuscript revision, evaluation-code inspection, and
figure-script preparation. This assistance is separate from the benchmark's
generation models and image judge and did not supply experimental results or
human ratings. The authors are responsible for the claims, citations,
artifacts, and experimental records.

\clearpage

\bibliographystyle{plainnat}
\bibliography{iclr2026_conference}

\clearpage
\appendix

\raggedbottom

\section{Industrial Domain Coverage}
\begin{table}[H]
\caption{Distribution across forty industrial domains. Counts combine the two tasks; domains with 210 requests contain ten taxonomy cases and fifty four-case families.}
\label{tab:appendix-domain-groups}
\centering
\small\fontsize{9}{11}\selectfont
\setlength{\tabcolsep}{3pt}
\begin{tabular}{@{}>{\raggedright\arraybackslash}p{0.39\linewidth}r >{\raggedright\arraybackslash}p{0.39\linewidth}r@{}}
\toprule
Industrial domain & $N$ & Industrial domain & $N$ \\
\midrule
Advanced filtration and membrane separation & 10 & Material handling and conveying & 210 \\
Agricultural machinery and farm processing & 10 & Medical and laboratory equipment & 10 \\
Compressed air, gas and vacuum systems & 10 & Mining and mineral-processing equipment & 10 \\
Automotive manufacturing and assembly & 10 & Mixing, agitation and mass transfer & 10 \\
Battery manufacturing and testing & 10 & Motors, gearboxes and transmission & 10 \\
Chemical process units & 10 & Oil and gas wellhead and gathering & 10 \\
Electrical enclosures and power distribution & 210 & Piping, manifolds and process skids & 10 \\
Energy and renewable-energy equipment & 10 & Plastics and rubber processing & 10 \\
Fire protection and emergency systems & 10 & Pneumatic actuators and air preparation & 10 \\
Foundry, die casting and casting finishing & 10 & Process vessels and separators & 210 \\
Grain, powder and bulk-food processing & 10 & Pulp, paper and tissue processing & 10 \\
Heat transfer and refrigeration equipment & 10 & Pumps and pump packages & 210 \\
HVAC and cooling systems & 210 & Rail transport equipment & 10 \\
Hydraulic power and actuation & 10 & Robotics and automation cells & 10 \\
Hygienic food, beverage and pharmaceutical equipment & 10 & Safety, access and support structures & 10 \\
Industrial fittings, joints and flexible connections & 10 & Semiconductor and cleanroom equipment & 10 \\
Instrumentation, sensors and inspection & 10 & Tanks, hoppers and silos & 10 \\
Linear motion and positioning & 10 & Valves and flow control & 10 \\
Machine tools and fabrication equipment & 10 & Warehouse and packaging logistics & 10 \\
Marine and port equipment & 10 & Water and wastewater treatment & 10 \\
\bottomrule
\end{tabular}
\end{table}

\section{Coverage and Provenance}
\label{app:source-provenance}
The source registry preserves 87 external source records. 
The inspected content included document mirrors, access challenges and application shells; it did not include native
CAD assets. 

The coverage audit identifies 2,000 taxonomy acceptance labels and 2,400
schema-only parameter names; neither supplies implicit numerical targets.
Canonical specifications contain no hidden numerical relation targets.

Reference assemblies support inspection of task structure and requested
parameter changes. They do not determine a unique target geometry or substitute
for model-output evaluation. Numerical validation of reference programs and
human agreement with the image judge are distinct forms of evidence.

\section{Representative Industrial Sources}
\begin{table}[H]
\hypersetup{hidelinks}
\caption{Representative industrial sources, previously inspected content and task-construction context. Sources are identified by provider and document title.}
\label{tab:source-examples}
\centering
\small\fontsize{9}{11}\selectfont
\setlength{\tabcolsep}{4pt}
\begin{tabular}{@{}>{\raggedright\arraybackslash}p{0.27\linewidth}>{\raggedright\arraybackslash}p{0.17\linewidth}>{\raggedright\arraybackslash}p{0.50\linewidth}@{}}
\toprule
Provider / document & Inspected type & Task-construction context \\
\midrule
\textbf{ABB: library portal} & HTML shell & Portal pointer; previously inspected application shell only. \\
\addlinespace[3pt]
\textbf{Alfa Laval: CAD portal} & HTML page & Hygienic-equipment portal pointer; no CAD asset obtained. \\
\addlinespace[3pt]
\textbf{Pall: filter CAD portal} & HTML page & Filter-model download page; no CAD asset obtained. \\
\addlinespace[3pt]
\textbf{Xylem: technical drawings} & HTML page & Pump/drawing index pointer; no CAD asset obtained. \\
\addlinespace[3pt]
\textbf{Siemens: drive configurator} & HTML page & Configurator description; motor/drive selection context. \\
\addlinespace[3pt]
\textbf{ADNOC: vessel specification} & Direct PDF & Owner specification; nozzle, support and access context. \\
\addlinespace[3pt]
\textbf{AutomationDirect: enclosure guide} & Direct PDF & Enclosure guide; doors, wiring and thermal-layout context. \\
\addlinespace[3pt]
\textbf{Grundfos: piping connections} & Mirrored PDF & Grundfos whitepaper via a secondary host; piping context. \\
\addlinespace[3pt]
\textbf{Flexco: conveyor handbook} & Direct PDF & Manufacturer handbook; cleaner, pulley and tensioner context. \\
\addlinespace[3pt]
\textbf{NSW Resources: belt tensioning} & Direct PDF & Government MDG 4 (1994); historical tensioning guidance. \\
\addlinespace[3pt]
\textbf{Swagelok: CAD resources} & HTML page & Fitting/drawing portal pointer; no CAD asset obtained. \\
\addlinespace[3pt]
\bottomrule
\end{tabular}
\end{table}

\section{Benchmark Access and Evaluation Protocol}
\label{app:execution}
\paragraph{Release scope.}
The benchmark release comprises the task data and evaluation resources needed to
compare models under a common protocol. Table~\ref{tab:release-scope} specifies
its five components. Industrial documents are identified by source names
(Table~\ref{tab:source-examples}).
Reference assemblies illustrate task structure rather than define a unique target
geometry: scoring tests the
requirements of each request and admits alternative valid realizations.

\begin{table}[h]
\caption{Benchmark release. All resources are versioned together to
identify the task set and evaluation conditions used in a comparison.}
\label{tab:release-scope}
\centering\small
\begin{tabular}{@{}p{0.24\linewidth}p{0.72\linewidth}@{}}
\toprule
Resource & Included content \\
\midrule
Task collection & All 1,400 requests, identifiers, domain/category/family assignments and evaluation specifications \\
Modeling interface & Common instructions, unit conventions and geometric primitive helpers \\
Evaluation definitions & Shared 0--100 visual-conformity standard, evidence rules and judge configuration \\
Rotation protocol & Five fixed panel lists, stratification rule and random seed \\
Evaluation software & Execution, standardized rendering, image assessment and score aggregation tools \\
\bottomrule
\end{tabular}
\end{table}

\paragraph{Model interface and execution.}
Each model receives the original request and common modeling instructions and
returns a geometry program. Programs execute in an empty, meter-scale Blender
scene using primitive helpers without assembly templates. The evaluator itself
does not edit source code, insert components or correct alignment. A single
enclosing code fence may be removed; ambiguous multi-block responses are not
treated as candidate programs. The evaluator produces
the saved scene and standardized images, with geometry
identity checked across execution, presentation setup and saving.

\paragraph{Bounded execution feedback.}
Each task allows one initial generation and at most one repair when the response
cannot form an executable program or the generated geometry fails execution or
validation. Feedback consists only of the original request, previous response
and observed error; it excludes hidden checks, scores and reference geometry.
The repair uses the same output-token limit. Generation attempts and transport recovery are recorded separately. Network or
authentication failures, low scores and evaluator rendering failures do not
trigger an execution-feedback repair.
Only the final valid delivery receives an image assessment. One bounded
transport/format recovery is allowed after an invalid attempt; the first valid
review is frozen and the budget persists across restarts. Interrupted requests
with an unknown outcome are not automatically sent again.

\paragraph{Standardized visual evidence.}
Fixed orthographic front, isometric and top cameras use EEVEE rendering at
1024-by-1024 resolution. The front camera lies on $-Y$, the isometric camera on
$+X/-Y/+Z$, and the top camera on $+Z$. Geometry is never rotated to improve a
view. Each view requires at least 0.5\% framing margin, at least
2\% projected bounding-box area, no border cropping, and valid foreground and
centering. Geometry projections must lie within the saved frames. The blind judge
receives only the request, images and rubric.

\section{Visual Assessment and Statistical Analysis}
\label{app:visual-review}
The primary metric is $S=GJ$, where $G$ indicates valid final delivery and
$J\in\{0,1,\ldots,100\}$ is a holistic integer visual score.
Table~\ref{tab:visual-rubric} provides descriptive reference anchors.
The same anchors apply to all requests, domains, and generators; no task-specific
criterion catalog or additive component score enters the metric. Component and geometric measurements are optional diagnostic extensions
with no contribution to the visual score.

A blind judge receives the request, front/isometric/top images, and public rubric.
It returns an explicit equipment-category match, one rating and summary, and
request-grounded matching observations, visible issues, external uncertainties,
and excluded observations. Category match records whether the requested equipment
identity is established; generic supports or covers alone do not establish it.
Every observation identifies an exact request excerpt, supplied view, location,
and visible evidence. Major/minor severity describes significance without fixed
penalties or score caps. The anchors do not restrict the available integers.
Schema, score range, and quoted request evidence are validated; the first valid
returned score is retained without automatic adjustment.

\begin{table}[t]
\caption{One common visual-conformity scale for every task. A valid delivered
output receives any integer from 0 to 100. These reference anchors are neither
fixed grades nor additive checks.}
\label{tab:visual-rubric}
\centering\small
\begin{tabular}{@{}p{0.08\linewidth}p{0.84\linewidth}@{}}
\toprule
$J$ & Reference anchor \\
\midrule
0 & No meaningful demonstrated fulfillment of the requested external assembly. \\
25 & Rough outline of the requested category; most core external structure is absent or wrong. \\
50 & Substantial partial match; a major visible issue or the main external arrangement remains unestablished. \\
75 & Mostly consistent main assembly; localized deviations or uncertainty concern secondary details. \\
100 & Consistent requested external assembly, without a substantive visible issue or unresolved external requirement. \\
\bottomrule
\end{tabular}
\end{table}

\paragraph{Stratified randomized rotation.}
Random seed defines a fixed random partition into five disjoint panels. Each
panel contains two taxonomy requests from each of 40 domains and ten complete
four-case families from each of five family domains: 280 requests per panel.
All models share the partition; five panels cover all 1,400 requests exactly once.
Family grouping preserves parameter comparisons. The panels characterize task
subset variation, rather than repeated-generation variability; camera directions
remain fixed throughout evaluation.

\paragraph{Human calibration.}
The completed study uses two independent reviewers and the shared 0--100 rubric.
For an output with both human ratings, their arithmetic mean is compared with
its GPT6-Astra score using mean absolute error and Spearman correlation. Comparisons
use ratings collected before adjudication and include only jointly assessed valid
outputs. Delivery failures remain in the request accounting but are excluded
from image agreement. The resulting summary statistics are reported in
Table~\ref{tab:human-model-calibration}.

\section{Measurement Conventions}
\label{app:measurement}
Extraction reads evaluated geometry, including beveled curves, and scene units.
Presence requires a semantic slot bound to nonempty geometry, without certifying
shape or connectivity. Lengths use fixed body targets in a Z-up frame; circular
sections use vertex geometry. Explicit counts include numeric topology options.
Unsupported or ambiguous decompositions remain unmeasured.

Conveyor dimensions use the carrying belt. Cabinet bounds exclude hoods and
nameplates; door angles use physical panel normals relative to the closed front
plane, excluding seals, frames, clearance envelopes and hardware. Cable-gland
counts exclude associated locknuts and fasteners. Motor length excludes feet. Manway diameters exclude
covers and flanges; nozzle counts exclude auxiliary flanges and fasteners.
Coil rows count distinct stacked levels, allowing transverse rows for a single-layer
parallel tube bank, independently of the requested count. Fan diameter uses
blade-vertex sweep around independently identified vertical hub axes. Cell counts
require complete solid divider panels across the casing interior. Guard height
uses each panel's or post's local upright span, including inclined conveyors.
Every relevant door, motor and rotor is checked; one correct instance cannot
conceal another incorrect one.

\clearpage
\ClearShipoutPictureBG
\begin{figure}[p]
\centering
\makebox[\linewidth][c]{\includegraphics[width=1.25\linewidth]{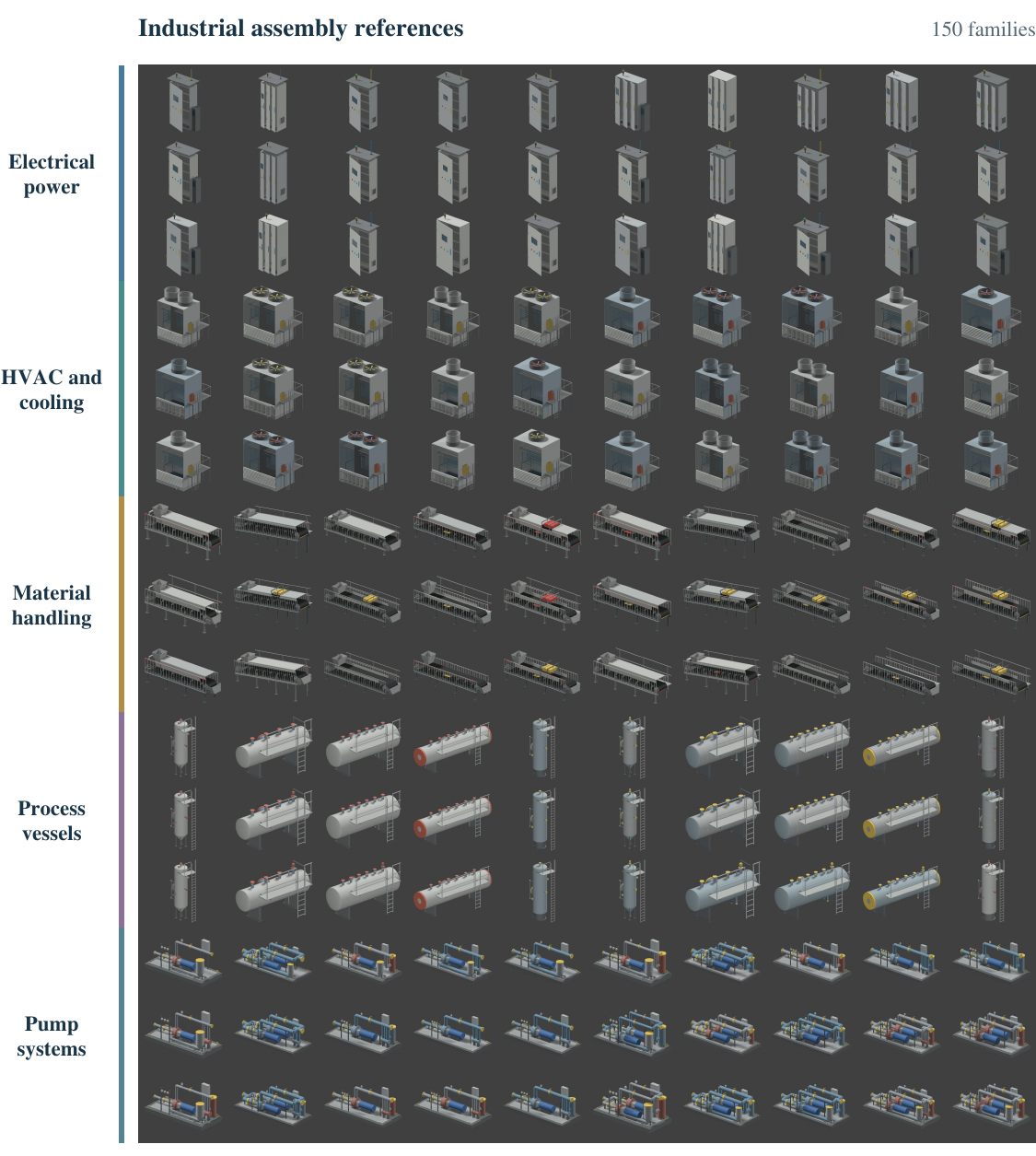}}
\caption{Reference assemblies from 150 design families across five industrial
domains. Each three-row band contains 30 families. The gallery covers 60\%
of the family track and illustrates its design diversity.}
\label{fig:reference-atlas}
\end{figure}
\clearpage

\end{document}